# Coral Model Generation from Single Images for Virtual Reality Applications *

Jie Fu [†]
University of the Arts London
Creative Computing Institute
London United Kingdom
j.fu1220161@arts.ac.uk

Shun Fu
Department Name
Bloks Technology Company
Shanghai China
xiantiaogames@outlook.com

Mick Grierson
University of the Arts London
Creative Computing Institute
London United Kingdom
m.grierson@arts.ac.uk

## ABSTRACT

With the rapid development of Virtual Reality (VR) technology, there is a growing demand for high-quality realistic 3D models. Traditional modelling methods struggle to meet the needs of large-scale customization, facing challenges in efficiency and quality. This paper introduces a deep-learning framework that generates high-precision realistic 3D coral models from a single image. The framework utilizes the Coral dataset to extract geometric and texture features, perform 3D reconstruction, and optimize design and material blending. The innovative 3D model optimization and polygon count control ensure shape accuracy, maximize detail retention, and flexibly output models of varying complexities, catering to high-quality rendering, real-time display, and interaction needs.

In this project, we have incorporated Explainable AI (XAI) to transform AI-generated models into interactive "artworks". The output can be seen more effectively in VR and XR than on a 2D screen, making it more explainable and easier to evaluate. This interdisciplinary exploration expands the expressiveness of XAI, enhances human-machine collaboration efficiency, and opens up new pathways for bridging the cognitive gap between AI and the public. Real-time feedback is integrated into VR interactions, allowing the system to display information such as coral species, habitat, and morphology as users explore and manipulate the coral model, enhancing model interpretability and interactivity. The generated models surpass traditional methods in detail representation, visual quality, and computational efficiency. This research provides an efficient and intelligent approach to 3D content creation for VR, potentially lowering production barriers, increasing productivity, and promoting the widespread application of VR. Additionally, incorporating explainable AI into the workflow provides new perspectives for understanding AI-generated visual content and advancing research in the interpretability of 3D vision.

## 1 Introduction

The rapid development of Virtual Reality (VR) technology has led to an increasing demand for high-quality realistic 3D models [1]. Traditional modelling methods face challenges in efficiency and quality [23], making it difficult to meet the requirements of large-scale customization [5]. Breakthroughs in deep learning within computer vision and graphics [7] have provided new approaches for intelligent 3D model generation [8]. However, existing methods still have limitations in efficiency, quality, and generalization capability. This paper proposes an innovative deep-learning framework that automatically generates high-precision realistic 3D coral models from a single image. The framework utilizes the Coral dataset to extract geometric and texture features, which are then used for 3D reconstruction with optimized design and material blending. The generated models surpass traditional methods in visual realism and detail representation.

Through quantitative evaluation of different AI generation algorithms and applications, this study demonstrates the working principles of AI vision systems and explores their capabilities and limitations in image understanding and content generation. This helps develop more explainable and controllable AI algorithms, providing insights for artists and designers to understand and utilize AI-generated technologies. This research aims to provide an efficient and intelligent method for creating immersive 3D content for VR applications. By using artificial intelligence to transform images into detailed 3D models automatically, this research promotes the application of VR technology and fosters innovation at the intersection of AI and the arts.

## 2 Related Works

Traditional 3D modelling methods rely heavily on professional software and manual operations, lacking automation [6]. Deep learning has opened up new avenues for data-driven 3D shape generation, enabling modelling from various inputs such as images, point clouds, and voxels [10]. Image-based methods use Convolutional Neural Networks (CNNs) and Generative Adversarial Networks (GANs) for 3D shape estimation [12] and texture generation [11], with some works combining the two to achieve end-to-end image-to-model conversion [16]. Point cloud methods directly process point [24-25] data to reconstruct models [13]. Implicit surface methods represent shapes as continuous functions, offering advantages in surface quality [17]. Neural rendering techniques combine graphics and deep learning to model from images to scenes [20], extending to dynamic [26] and editable scenes [21]. This paper focuses on generating detailed and realistic 3D models from limited data for immersive applications, exploring precision, efficiency, generalisation, editability, interactivity, and practical application verification.

## 2.1 Data Preprocessing

We utilised the Coral dataset from CoralNet [22], which contains a large number of coral images of various species and morphologies. Backgrounds were automatically removed using remove.bg [22], followed by batch processing in Photoshop to adjust colours and shadows. For low-resolution images, we enhanced details using super-resolution algorithms like Gigapixel AI [23]. The pre-processed dataset provides a high-quality, consistent foundation for feature extraction and 3D reconstruction.

## 3 Implementations

### 3.1 Evaluation of Single Image to 3D Model

We designed a test experiment using quantitative and qualitative metrics, including usability, geometric accuracy, visual realism, processing time, controllability, and generation effect. These were evaluated in contexts such as texture blending, local detail performance, and noise resistance, and compared against commercial services, open-source projects, and parametric modelling plugins. Ten experts from the fields of computer graphics, virtual reality, and visual design, who have extensive experience in 3D modelling, VR development, and interaction design, were invited to rate the 3D coral models generated by different methods. The rating was done using a 1-5 Likert scale with 1 indicating very dissatisfied and 5 indicating very satisfied. We performed statistical analysis on the rating results for each metric to obtain the average scores for each method, which can be seen in Table 1. Although the rating process inherently involves some subjectivity and individual expert opinions may vary, we mitigated personal bias by aggregating scores from multiple experts and conducting statistical analysis to achieve more objective and robust evaluation results. Nonetheless, the variability in ratings should be considered when interpreting the evaluation outcomes.

### 3.2 Feature Extraction and 3D Model Generation

We selected high-resolution coral images to extract rich image features and used TripoSR [9] for initial 3D reconstruction, obtaining a rough initial shape. Our deep learning framework introduces innovative designs in feature extraction and 3D reconstruction. The framework captures detailed visual information from a single image and directly outputs high-quality mesh models through end-to-end feature fusion and 3D inference. We utilised Convolutional Neural Networks (CNNs) to extract multi-scale feature maps, visualising activation maps from different layers to intuitively understand how the network extracts coral shapes, textures, and other visual elements. Generative Adversarial Networks (GANs) were employed to transform image features into 3D shapes and texture representations, comparing the generated models with ground truth to visualise the evolution of shapes and textures during GAN training and quantitatively evaluating reconstruction accuracy to gain insights into the generation mechanism. We introduced a physics-based shape optimisation module, simulating fluid dynamics and collision constraints during coral growth to refine the initially generated models. By visualising model changes before and after optimisation and examining the impact of different simulation parameters on the results, we explained how the algorithm uses prior knowledge to enhance generation quality. We developed an interactive model editing tool, allowing users to artistically edit and customise the generated models through brushes and parameter adjustments, expanding the diversity of algorithm-generated content and providing new perspectives for understanding user intentions and preferences in the generation process.

| Method | Usability | Geometric Accuracy | Visual Realism | Average Processing Time | Controllability | Generation Result | Overall Score |
|--------|-----------|--------------------|----------------|-------------------------|-----------------|-------------------|---------------|
| TripoSR | 4.9 | 4.8 | 4.8 | 4.6 | 4.9 | 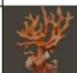 | 24 |
| InfiniGen | 3.2 | 5.0 | 5.0 | 4.0 | 3.6 | 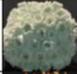 | 20.8 |
| Meshy.ai | 4.7 | 4.9 | 4.8 | 4.2 | 4.8 | 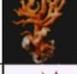 | 23.4 |
| CSM.ai | 4.5 | 3.8 | 4.0 | 4.0 | 4.2 | 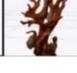 | 20.3 |

**Figure 1: Performance comparison of different 3D generation methods across various evaluation metrics**

### 3.2 Results and Analysis

TripoSR and Meshy.ai [18 ]excel in geometric accuracy and visual realism. InfiniGen [9]scores highest in usability and generation effect, but its controllability and stability need improvement. CSM.ai [4] performs well overall but lacks geometric accuracy and visual realism. Different methods have strengths in different application scenarios. For batch generation needs, InfiniGen is more efficient; for scientific visualisation requiring high precision and realism, TripoSR and Meshy.ai are more suitable; CSM.ai's [4]usability and controllability make it ideal for creative designs requiring strong interactivity.

## 4 Model Optimization and Polygon Count control

To further enhance realism and meet the complexity requirements of different applications, we adopted a progressive model optimisation and polygon count control strategy. We imported high-resolution meshes into ZBrush [21] for retopology, using ZRemesher to convert triangular faces into regular quadrilateral meshes, preparing them for subsequent subdivision and sculpting operations. By applying subdivision surface techniques, we enhanced surface smoothness while maintaining shape features and used sculpting tools to fine-detail local textures and holes.

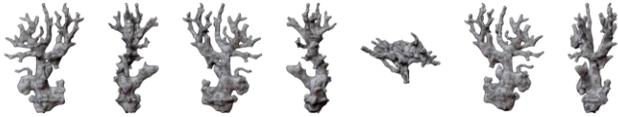

**Figure 1: Trip o SR AI-Generated Model Diagram**

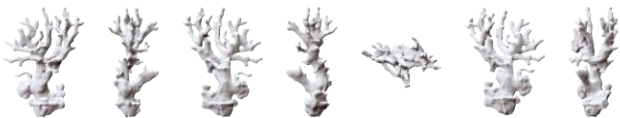

**Figure 2: Using DynaMesh and Polishing to Smooth Rough Surfaces**

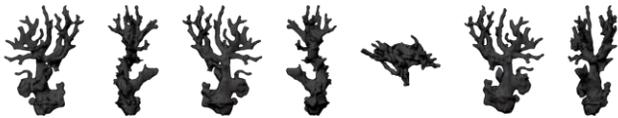

**Figure 3: Display of Polygon Count After Polishing, Reaching an Impressive 500,000**

UV Master automatically unwrapped the UVs (UV Mapping), transferring high-resolution details to the low-resolution model. Decimation Master simplified the polygon count to below 100,000, baking high-resolution details into normal maps to reduce rendering costs while preserving visual quality. The noise was added to enhance surface details.

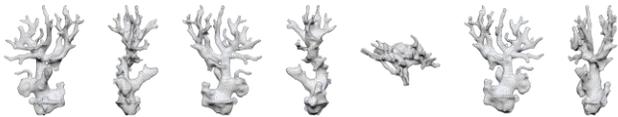

**Figure 4: Display of Polygon Count After Polishing**

Reaching an Impressive around 320,000 The Polygon Count Was Reduced to 52,000, Significantly Lowering Model Complexity.

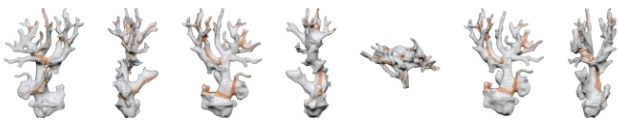

**Figure 5: Unwrap the UVs of the Low-Resolution Model**

Using UV Master to Automatically Unwrap the UVs of the Low-Resolution Model, Optimizing the UV Layout for Subsequent Texture Baking

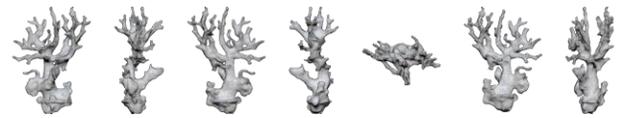

**Figure 6: Noise texture adds details to the high-poly model.**

Adding Details to the High-Resolution Model Using Noise to Enhance Surface Texture. The Processed High-Resolution Model Reaches a Polygon Count of 1,500,000. The noise texture adds rich surface details to the high-resolution model, making it more realistic. For creating PBR materials, we used Substance Painter to produce multi-channel maps, including albedo, roughness, and metallic maps, with detailed depictions of coral appearances based on real coral photos and biological knowledge. To meet high-performance application requirements, we further simplified the model to approximately 10,000 faces using Simplygon and compressed the file to megabyte levels using Draco, achieving fast model transmission and loading.

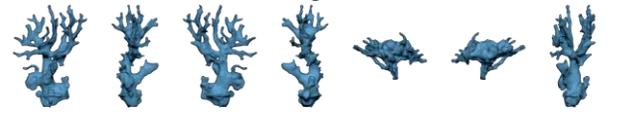

**Figure 7: Noise texture adds details to the high-poly model**

Through a systematic optimisation process, we can convert high-resolution models into low-resolution models while maintaining high-quality rendering by baking normal maps. This approach meets the needs of various application scenarios, such as real-time display and interaction, greatly enhancing the method's practicality.

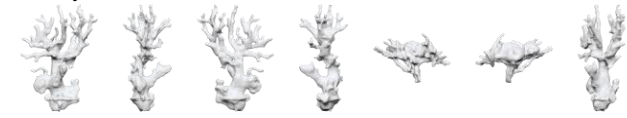

**Figure 8: Model Generated in WebGL and Its Presentation in VR Space**

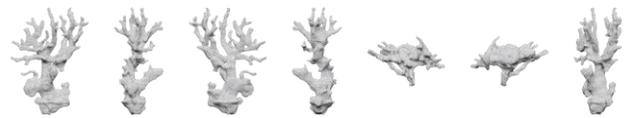

**Figure 9: Adding details to high-resolution models through noise perturbation.**

We developed a WebGL-based material editor and renderer to achieve high-quality display of coral models on a lightweight web platform. The models can also be imported into Vision Pro, supporting advanced rendering features such as 3D model lighting, physical transparency, and detailed viewing through rotation. These features are used for underwater environment simulation, providing an immersive experience (Figure 9).

 

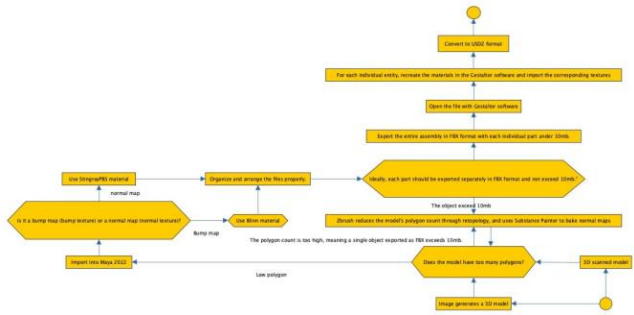

**Figure 10: Progressive model optimisation and polygon count control**

To better understand the model's texture mapping process, we integrated explainable AI techniques into the framework. These techniques provided insights into how the model applied textures from the original 2D images to the generated 3D models. Visualisation tools demonstrate how texture features from the original image are selected and applied to different regions of the 3-D model, allowing users to intuitively understand the correspondence between texture features and the 3D model, explaining the texture mapping strategy. Through this optimisation process, we can flexibly output models of varying complexity for different use cases like high-quality rendering, printing, real-time display, and interaction, greatly expanding practicality.

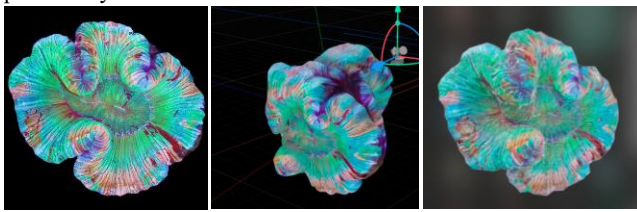

**Figure 11: Coral Image - Tripo SR 3D Generated Model - Manually Adjusted Coral Texture and Details**

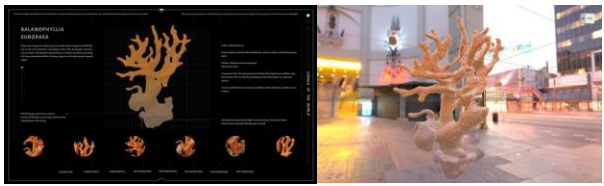

**Figure 12: Impact on AI, 3D Modeling, and VR Fields**

## 5 Impacts On Ai, 3d Modelling, And Vr Fields

Our research demonstrates the significant potential of artificial intelligence technology in the fields of 3D modelling and VR content creation. Traditional 3D modelling processes rely heavily on manual operations by professional designers and artists, requiring substantial time and human resources. The single-image 3D model generation method based on our proposed deep learning can automatically extract semantic and geometric information from images and generate high-quality 3D models, greatly simplifying the modelling process and lowering the technical barriers to content production.

This breakthrough is expected to significantly promote the adoption and popularisation of VR technology across various fields. Our algorithm allows non-professionals to quickly create realistic virtual scenes and objects without mastering complex 3D modelling skills. This will drastically reduce the production costs of VR content, shorten development cycles, and make it easier for fields such as education, training, entertainment, healthcare, and tourism to utilise VR technology, providing users with immersive experiences.

For example, in the field of education, teachers can use our method to quickly create various realistic virtual teaching scenes and models, such as historical sites, natural landscapes, and human anatomy, allowing students to explore and learn in VR environments, thereby enhancing teaching effectiveness. In the tourism sector, our technology can generate high-fidelity models of iconic buildings and attractions worldwide, enabling users to enjoy the beauty of global landmarks from the comfort of their homes, offering an immersive virtual tourism experience.

Furthermore, our research provides new insights into the further development of artificial intelligence in the 3D vision domain. Our work proves that deep learning methods can efficiently learn and extract shape and texture information from 2D images and generate high-fidelity 3D representations [26]. This lays the foundation for exploring AI applications in 3D reconstruction, semantic understanding, and scene generation, potentially advancing 3D modelling and VR content creation to higher levels of automation and intelligence.

We plan to extend this method to more complex and diverse object categories, further enhancing its generalisation capability and robustness. Additionally, we are exploring end-to-end 3D scene generation and interaction control technologies to achieve automatic generation and real-time interaction from images to complete virtual scenes. This will bring more possibilities to immersive applications such as VR, greatly enriching the content of the virtual world.

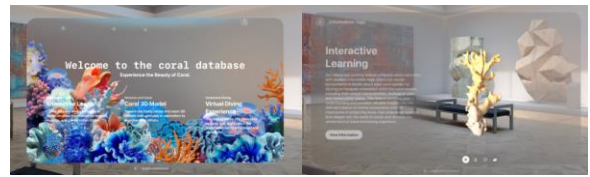

**Figure 13: Impact on AI, 3D Modelling, and VR Fields**

## 6 Conclusions

This paper presents an innovative deep-learning framework that automatically generates high-precision, realistic 3D coral models from a single image. The method introduces novel designs in 3D model optimisation and polygon count control, allowing for flexible output of models with varying complexity to meet diverse application needs. To explore explainable AI applications in XR, we employed models to generate 3D representations from 2D images, assessed their quality, and devised an improved evaluation methodology for XR. Our approach utilised visualisation tools to illustrate texture mapping techniques, enabling users to intuitively grasp how original image textures are mapped onto different areas of the 3D model. By providing transparency in the system's construction process, we aimed to enhance the trustworthiness of the generated models. This research highlights the significant potential of AI in 3D modelling and VR content creation. The automated, intelligent 3D content generation method can significantly lower the development barriers for VR applications. In the future, we aim to extend this method to a wider variety of objects, enhance its generalisation capabilities, and explore end-to-end 3D scene generation and real-time interaction control techniques to enrich the content of virtual worlds.